\documentclass[10pt, letterpaper, copyright]{krafton-ai}

\usepackage{amsmath}
\usepackage{amssymb}
\usepackage{booktabs}
\usepackage{multirow}
\usepackage{float}
\usepackage{graphicx}
\usepackage[authoryear, round]{natbib}
\usepackage[useregional=false]{datetime2}
\DTMsetstyle{iso}

\newcommand{\ludi}{\texorpdfstring{Ludi\textsubscript{0.1}}{Ludi0.1}}
\newcommand{\ludione}{Ludi 1.0}
\newcommand{\say}[1]{``#1''}

\renewenvironment{quote}
  {\list{}{%
    \setlength{\leftmargin}{2.5em}%
    \setlength{\rightmargin}{2.5em}%
    \setlength{\topsep}{0pt}%
    \setlength{\partopsep}{0pt}%
    \setlength{\parsep}{0pt}%
    \setlength{\itemsep}{0pt}%
  }\item\relax}
  {\endlist}

\newtcbox{\tooltag}{
  on line,
  arc=2pt,
  boxrule=0.3pt,
  colframe=black!30,
  colback=black!6,
  fontupper=\small\ttfamily,
  left=2pt,
  right=2pt,
  top=0.5pt,
  bottom=0.5pt,
  boxsep=0pt
}

\title{\ludi{}: An Agentic System for Socially Intelligent Robots}
\renewcommand{\thefootnote}{\textdagger}
\author{Ludo Robotics$^{\dagger}$}
\renewcommand{\thefootnote}{\arabic{footnote}}  
\renewcommand{\titlefont}{\color{KraftonBlack}\normalfont\sffamily\bfseries\LARGE}
\makeatletter
\renewcommand{\maketitle}{\bgroup\setlength{\parindent}{0pt}
  \begin{adjustwidth}{0pt}{24pt}
    \begin{flushleft}
      {
        {\raggedright \titlefont \@title\par}%
        \vskip11pt
        {\centering \@author\par}%
        \vskip20pt%
      }%
    \end{flushleft}
  \end{adjustwidth}
  \let\thefootnote\relax
  \footnotetext{$^{\dagger}$The complete list of authors is in the Authorship section at the end of this report.}
  \egroup
  {\abscontent}
  \thispagestyle{firststyle}
}
\makeatother

\paperdate{\DTMtoday}

\begin{abstract}
Robot foundation models have substantially advanced perception
and control, but natural human--robot collaboration requires more than
executing isolated commands. A robot must recognize ambiguity, maintain
context across turns, communicate its intentions, and revise ongoing
behavior as the user's intent changes. We present \textbf{\ludi{}}, an agentic system for socially intelligent 
robots that integrates interactive speech, multimodal reasoning, memory,
navigation, and learned manipulation. Its decision-making core is a fine-tuned vision--language model trained on multi-turn
interaction traces spanning ambiguous requests, clarifications, corrections,
interruptions, mixed social and task dialogue, and multi-step tasks.
A purpose-built harness manages the model--tool interaction loop, while
specialized navigation and  manipulation policies execute
physical skills. \ludi{} demonstrates a practical path toward fluid human--robot
collaboration today while producing the multimodal interaction traces
needed to develop a more deeply integrated foundation model for robots
and people.
\end{abstract}

\makeatletter
\renewcommand{\abscontent}{
  \begin{tcolorbox}[
    enhanced,
    frame hidden,
    colback=KraftonLightGray,
    arc=4pt,
    left=12pt, right=12pt, top=12pt, bottom=12pt,
    before skip=0pt, after skip=0pt
  ]
  {\absfont
    \theabstract
    \vskip0.8em
    \noindent\textbf{Videos and additional examples:} \href{https://www.ludorobotics.ai/research/ludi-0-1}{ludorobotics.ai}
  }
  \end{tcolorbox}
}
\makeatother

\begin{document}
\maketitle


\vspace{-.025in}

\section{Introduction}

Recent advances in robot foundation models have substantially expanded the range of tasks robots can perform. Models built on pretrained large language and vision--language models exhibit increasingly general perception, reasoning, and language-conditioned control across navigation and manipulation \citep{driess2023palme,zitkovich2023rt2,kim2025openvla,black2024pi0,bjorck2025groot,cheng2024navila}, while more recent approaches have begun to incorporate generative world models to support prediction and action generation \citep{zhen2024threedvla,dreamzero2026,zhang2026dreamvla}. Much of this progress, however, has emphasized physical task competence and generalization. Natural human--robot collaboration requires capabilities beyond physical competence: robots must also interpret human intent, communicate naturally, and adapt their behavior through speech, gesture, and shared context.

\begin{figure}[H]
    \centering
    \includegraphics[width=0.95\linewidth]{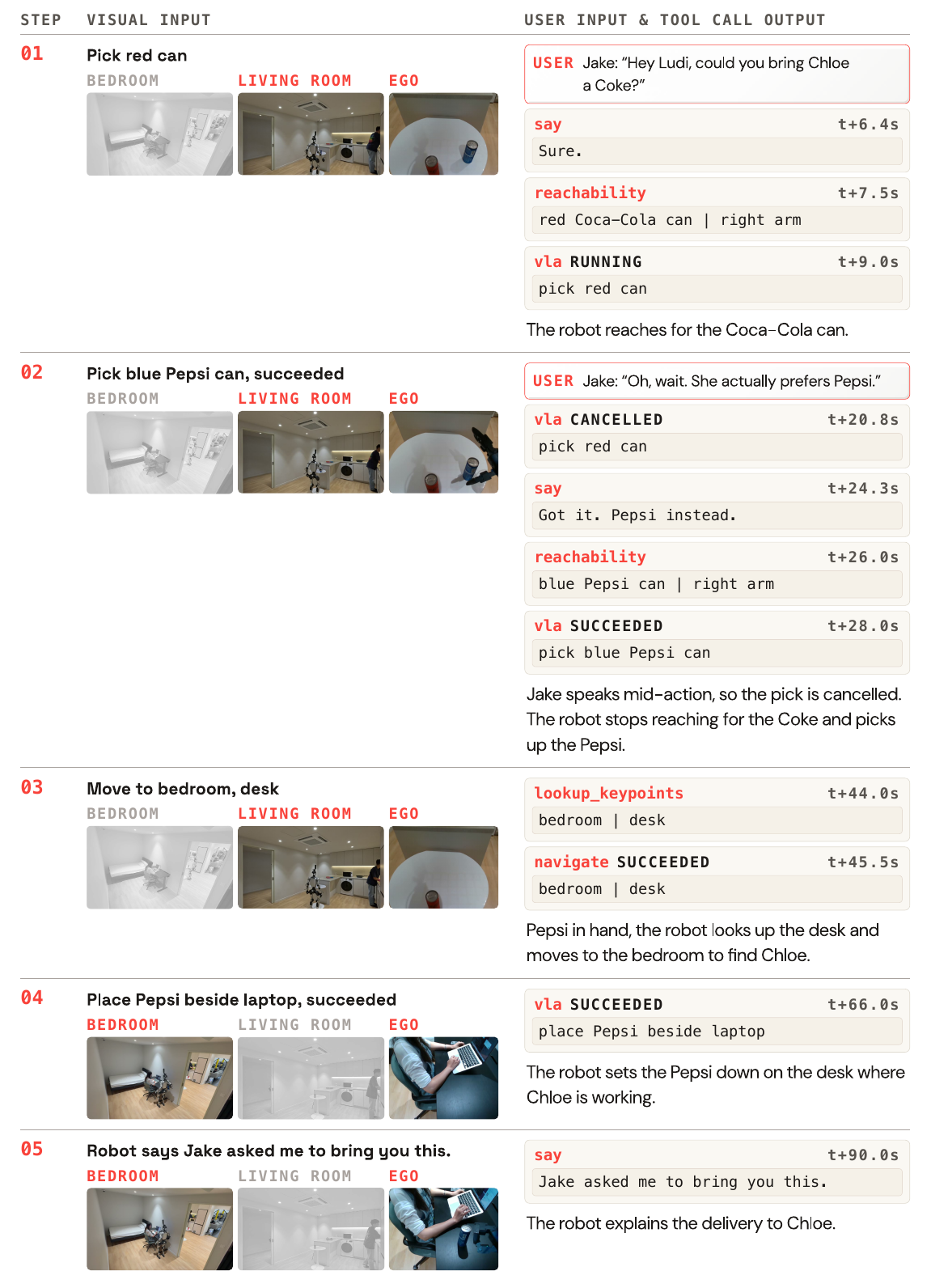}
 \caption{\textbf{Social intelligence in action.}
In this real-robot demonstration on a Unitree G1 humanoid, a user asks \ludi{} to bring a Coke to Chloe in the bedroom, and \ludi{} begins reaching for the Coke. When the user remembers that Chloe prefers Pepsi and changes the request, the robot abandons the original action, retrieves the Pepsi, navigates to the bedroom, and delivers it to Chloe's desk. The episode illustrates how dialogue, contextual reasoning, navigation, and manipulation are coordinated within a single evolving human--robot interaction.}
    \label{fig:ludi_demo}
\end{figure}

Consider a household task performed by a Unitree G1 humanoid robot, illustrated in Figure~\ref{fig:ludi_demo}. A person asks the robot:

\begin{quote}
``Hey Ludi, could you bring Chloe a Coke?''
\end{quote}




Two soda cans sit on the table, one Coke and one Pepsi. The request names which one, so the robot says ``Sure,'' and checks whether the red can is within arm's reach and reaches out to pick it up. 

A challenge arises when the user's intent changes after the robot
has already begun acting. Suppose the user initially chooses Coke,
and \ludi{} begins reaching for the Coke. The user then remembers that
Chloe prefers Pepsi and says:

\begin{quote}
``Oh, wait. She actually prefers Pepsi.''
\end{quote}

\noindent \ludi{} must understand that this correction supersedes the earlier
choice, relate it to the action already underway, and revise its behavior
accordingly. It abandons the reach for the Coke, picks up the Pepsi,
navigates to the bedroom, and delivers it to Chloe's desk.

This example illustrates a key form of social intelligence that is natural for people but difficult for robots: adapting quickly when a user’s intent or instruction changes after action has begun. Doing so requires speech, perception, memory, reasoning, navigation, and manipulation to operate as an integrated system, allowing new information to reshape ongoing behavior. Despite recent progress in robotics, making such fluid, responsive interaction reliable across varied environments and tasks remains an open challenge. This work takes a step toward that goal by integrating social interaction and physical behavior within a single robotic system.

\section{\ludi: An Agentic Approach}

\ludi{} integrates robot perception, navigation, and manipulation with interactive speech, dialogue, memory, and social reasoning in a functioning robotic system. Our current implementation runs on a Unitree G1 humanoid.
We deliberately chose an agentic architecture rather than relying on a single end-to-end model because natural human–robot interaction requires the robot to coordinate specialized capabilities over time and revise ongoing behavior as new information arrives. 

Modern vision--language models (VLMs) jointly represent visual and linguistic information and support rich visual--language understanding and open-ended multimodal generation \citep{radford2021clip,alayrac2022flamingo,liu2023llava}. These capabilities make VLMs a natural candidate for the reasoning core of an interactive robot. At the center of \ludi{} is a fine-tuned VLM that serves as the reasoning core of the agent. It interprets the current scene and interaction history and decides what the robot should say or do next, including whether to wait, navigate, or invoke a manipulation policy.

\ludi{}  operates within a harness that manages the interaction loop. The harness assembles the current context, calls the VLM, validates and executes its selected tool, records the result, and calls the VLM again with the updated context. A single user request can therefore produce several cycles of reasoning, tool use, and observation before the interaction returns to the user. The VLM decides what should happen next; the harness manages how that decision is executed and incorporated into the ongoing interaction.

Spoken input triggers a new agent turn, while speech output runs asynchronously so that reasoning and action can continue while the robot is speaking. New user input can arrive at any time and redirect the robot even while it is speaking or acting.
The harness maintains a shared interaction record containing dialogue, visual observations, model decisions, tool calls, and outcomes. The VLM reasons over this record and generates new plans, responses, and action decisions. As the interaction grows, an auxiliary model compacts older context in the background without disrupting an active turn.
All inference and system operation can run locally, without relying on external model APIs or cloud-based inference. In our experiments, \ludi{} ran on an onsite GPU workstation connected to a Unitree G1 robot.

\begin{figure}[t]
  \centering
  \includegraphics[width=1.0\textwidth]{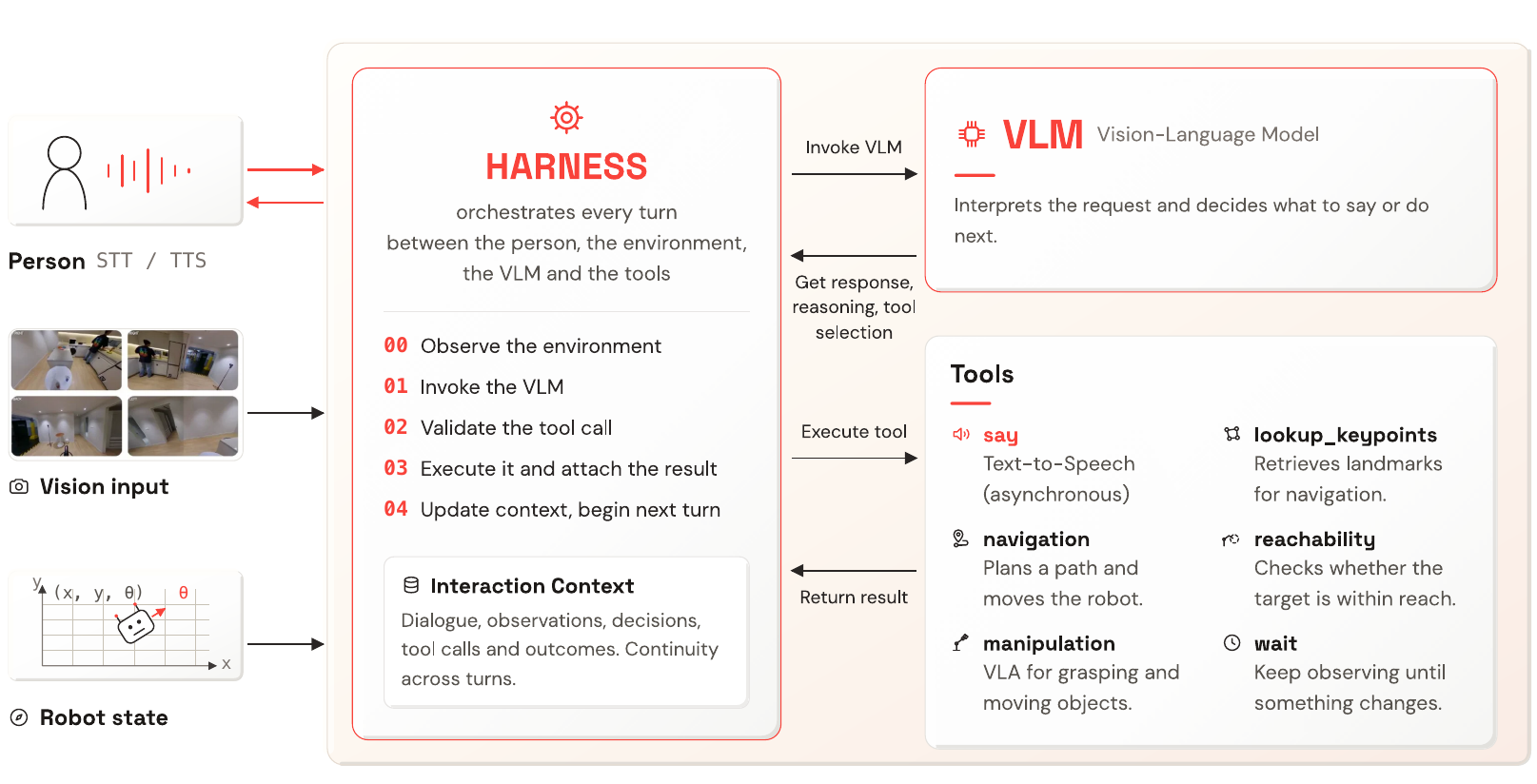}
  \caption{Architecture of the \ludi{} embodied agent. The VLM provides scene perception, reasoning, dialogue, and tool selection, while the harness manages model calls, tool execution, results, and shared context across turns.}
  \label{fig:agentic-loop}
\end{figure}

We refer to the complete embodied agent as \ludi{}: the VLM together with the harness, speech components, navigation system, and manipulation tools, depicted in Figure~\ref{fig:agentic-loop}. For manipulation, \ludi{} invokes specialized vision--language--action (VLA) policies, a class of models that conditions on visual observations and language instructions to generate robot actions \citep{zitkovich2023rt2,kim2025openvla}. In our current implementation, these manipulation policies are based on GR00T N1.7 \citep{bjorck2025groot}. Within its current set of supported environments and skills, \ludi{} can engage in spoken interaction, interpret the surrounding scene, ask clarifying questions, navigate indoors, and carry out manipulation tasks using these policies. It can maintain context across an extended interaction and revise an ongoing plan when the user provides new information.

Together, these capabilities allow \ludi{} to handle interactions that cannot be reduced to isolated perception, dialogue, navigation, or control tasks. The robot can move fluidly between understanding a request, gathering missing information, explaining its intent, acting in the physical world, and revising its behavior in response to the user. The goal is not simply to make robots more conversational. Speech must remain grounded in what the robot sees, remembers, plans, and does, so that communication and physical action unfold as one coherent interaction in the real world.

\section{A VLM for Interactive Robotics}

At the center of \ludi{} is a Qwen3.5 vision--language model (VLM) \citep{qwen35}, fine-tuned specifically to serve as the reasoning core of the agent by perceiving the scene, interpreting and responding to user interactions, and deciding which tool to invoke at each step. Starting from the pretrained model, we fine-tuned it on synthetic multi-turn demonstrations of complex tasks and human--robot interactions.
Our objective was to go beyond visual recognition and instruction following, training the model to interpret an evolving interaction, reason over shared context, select among speech and physical tools, and revise its behavior when the user provides new information.

\subsection{The Agent Interface}
At each turn, the model receives the robot's ego-view image, panoramic context from a head-mounted 360\textdegree{} camera, the transcribed user speech, the interaction history including its own reasoning, and the results returned by previous tool calls. It responds with a single action drawn from a small, fixed tool set:
\begin{itemize}
    \item \tooltag{say}: speak to the user, including responding to the user and asking a clarification question.
    \item \tooltag{lookup\_keypoints}: retrieve the static map of rooms and named locations, sorted by distance from the robot.
    \item \tooltag{navigate}: move to a named location in the environment.
    \item \tooltag{reachability}: check whether a visible object can be picked up, and which arm to use, before attempting manipulation.
    \item \tooltag{vla}: execute a manipulation, including picking up or placing an object via a trained VLA policy.
    \item \tooltag{wait}: hold and keep observing until something changes.
\end{itemize}


Each tool is exposed to the model through a typed function-call schema, a structured interface that specifies the tool’s arguments and their allowed values. The harness enforces the contract around each call: a turn may contain at most one action-bearing tool call, a pick must be directly preceded by a successful \tooltag{reachability} check, and malformed or out-of-contract calls are rejected rather than executed.  Parts of these schemas are populated at runtime. The \texttt{object\_type} vocabulary of \tooltag{reachability} and \tooltag{vla} is read from the robot's VLA registry when the agent starts, so newly trained manipulation skills automatically appear to the model as new argument values. This allows the set of manipulation skills available to the agent to expand without retraining the agent itself.

\subsection{Training Data and Fine-Tuning}

We built the training corpus by generating interaction scenarios combinatorially over tasks, objects, and interaction events. Each scenario was rendered into a multi-turn interaction trajectory and serialized as a tool-call trace with intermediate reasoning. The corpus covers the situations that make interactive robotics hard: ambiguous requests, mid-task corrections, interruptions, multi-step errands, and dialogue that mixes task-oriented conversation with social interaction. The training corpus consisted of 14,432 trajectories.

\begin{figure}[htbp]
  \centering
  \includegraphics[width=\linewidth]{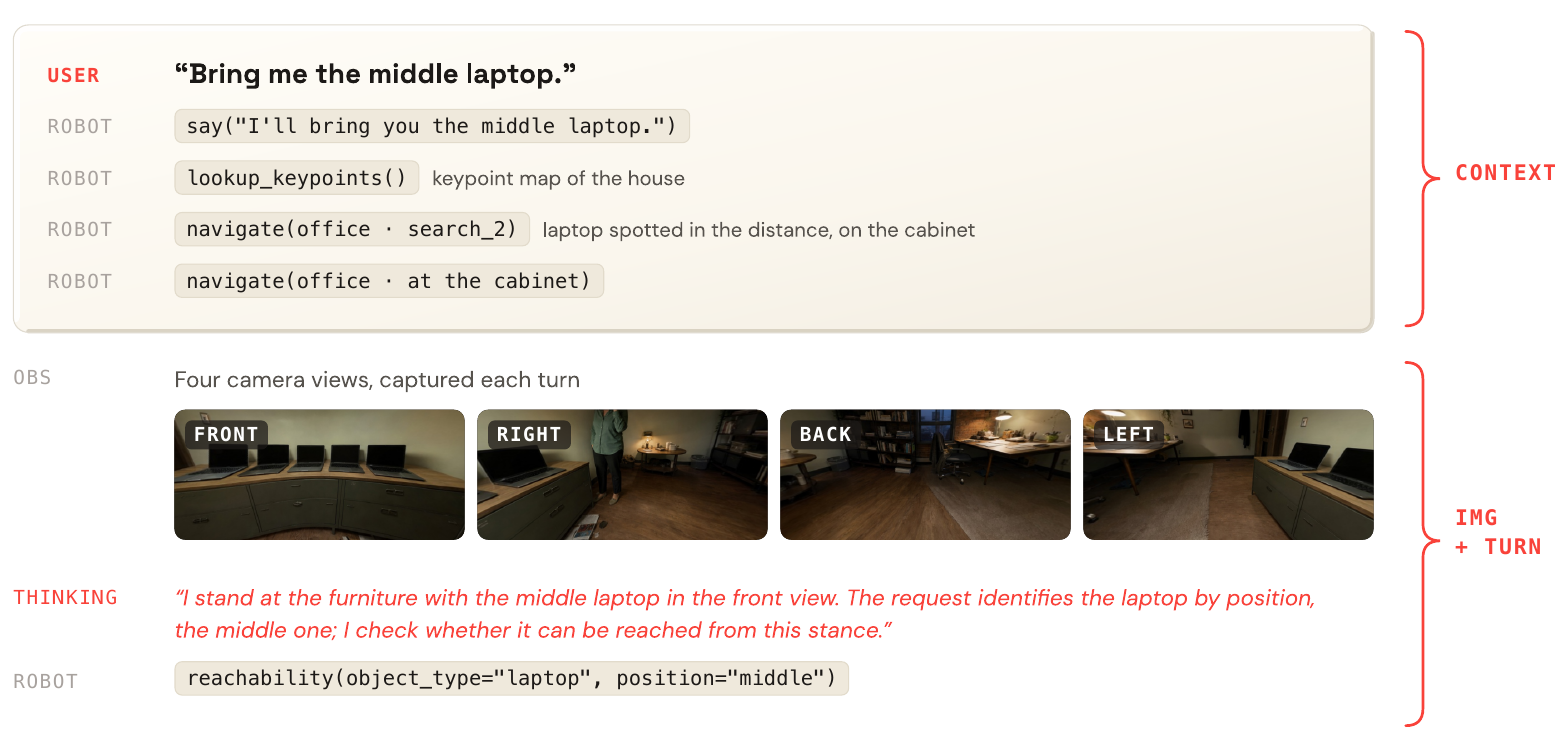}
  \caption{\textbf{An example of generated data.}
  The user identifies the target by spatial position (\say{the \emph{middle} laptop}); the model grounds the target in the four-view observation before issuing a reachability check. The multi-turn tool history (\textsc{context}) and the latest four-view observation form the input, while the reasoning and one action (\textsc{img~+~turn}) are the supervision target.}
  \label{fig:data-examples}
\end{figure}

The demonstrations were fully synthetic, with tool-call correctness guaranteed by construction: a state machine and policy table determined every tool call, its arguments, and its result.
A teacher LLM wrote each turn's language, including the reasoning text, spoken responses, and user utterances, controlling vocabulary and style. Every authored turn then passed through a quality gate of automated rule-based checks and a two-stage LLM judge. If a turn failed these checks, the system allowed up to two constrained revision attempts; turns that still failed were rejected. Scene imagery comes from generated images and simulator renders of environments that mirror the deployment space. Figure~\ref{fig:data-examples} shows a generated demonstration.

We explored several adaptation strategies, including distillation and prompt optimization, and ultimately used supervised fine-tuning on the interaction traces. We fine-tuned Qwen3.5 at the 4B and 9B scales \citep{qwen35}. The fine-tuned 9B model achieved higher benchmark success than the 4B model, but we deployed the smaller model because its lower inference latency provided a better tradeoff for real-time interaction.

\subsection{Closed-Loop Simulation for Interactive Robot Agents}

Existing VLM benchmarks typically score isolated predictions: given an image and a question, compare the model's answer against a reference answer or predefined choice \citep{goyal2017vqa,hudson2019gqa,yue2024mmmu}. That misses what an embodied agent actually has to do, which is to sustain a multi-turn interaction with people while acting in a world that changes because of its own actions. We therefore built a closed-loop simulation environment in which the agent serves a user end-to-end: it perceives, speaks, navigates, and manipulates through exactly the same tool interface and system prompt as on the real robot. The only scripted component is the simulated user; the agent's behavior and the resulting evolution of the environment unfold in closed loop.

We run these closed-loop interactions in AI2-THOR \citep{ai2thor}, an interactive indoor simulator, using ProcTHOR \citep{deitke2022} to generate diverse 3D homes with different room layouts, furnishings, and object placements. Each generated home is fixed by hash so that every scenario can be reproduced exactly. The simulated robot matches the deployed system: the agent receives the same four panoramic views using the same calibrated camera geometry as the robot’s head-mounted 360\textdegree{} camera, navigates only to named keypoints on a static map, and manipulates through the same pick-and-place interface.

\subsection{Evaluation}

We designed Interaction Core28, a closed-loop benchmark in AI2-THOR consisting of 28 scenarios that cover delivery, perception and reporting, ambiguity that requires clarification, robustness to failed searches and unreachable objects, and mid-task interaction. Each scenario defines a gold tool-call trajectory, and every episode is scored on both its interaction and its final simulator state. Core28 is our primary measure because it exercises the full loop of perception, dialogue, and action.

An episode counts as a task success when it terminates cleanly, satisfies the scenario's deterministic interaction gates, and reaches the required simulator end state. The gates encode the interaction behavior being tested, such as asking for clarification before acting under ambiguity or reporting honestly when a target cannot be reached.

Because deterministic gates cannot reliably account for alternative phrasings, a failed episode is passed to an LLM judge when the sole failing gate concerns the semantic content of a spoken utterance. The judge examines the full transcript, staged ground truth, and scenario success criterion and may overturn the failure to a pass. It cannot overturn failures involving tool use, task execution, or the final simulator state.

Table~\ref{tab:core28} reports episode-level task success on Core28. Fine-tuning roughly doubles success over the same pretrained model, and the fine-tuned 9B completes 20 of 28 scenarios end-to-end, approaching a frontier model (GPT-5.6) with an optimized prompt. Remaining failures concentrate in spatial grounding at the delivery end of a task: the agent picks the right object and plan but localizes the hand-off point imprecisely.

\begin{table}[h]
\centering
\caption{Interaction Core28 closed-loop scenario benchmark (simulation).
Episode-level task success after LLM-judge appeal; the parenthesized count is
how many of those successes the judge granted by overturning a
language-semantic gate failure.
Fine-tuned models use LoRA \citep{hu2022lora}  applied to the corresponding Qwen3.5 base models.}
\label{tab:core28}
\begin{tabular}{lc}
\toprule
Model & Success (judge-overturned) \\
\midrule
GPT-5.6 (GEPA-optimized prompt;~\cite{agrawal2026gepa}) & 26/28 (0) \\
\midrule
Qwen3.5-4B (vanilla)            & 10/28 (1) \\
Qwen3.5-4B (fine-tuned)         & \textbf{17/28} (1) \\
Qwen3.5-9B (vanilla)            & 11/28 (0) \\
Qwen3.5-9B (fine-tuned)         & \textbf{20/28} (1) \\
\bottomrule
\end{tabular}
\end{table}

\section{Engineering the Harness for Real-Time Interaction}

\ludi{}'s control harness is designed to keep the interaction responsive and reliable even when model inference, context management, speech, and physical actions operate at different timescales. A few design choices are particularly important:

\begin{itemize}
  \item \textbf{Non-blocking context compaction:} As the interaction history grows, an auxiliary model summarizes older context in the background. The compacted history is installed only between active turns, so context management never interrupts or changes the state of a running agent loop.


  \item \textbf{Visual prefill and asynchronous speech:} A background process continually refreshes the camera input and performs visual prefill before a user instruction arrives, reducing the delay before the VLM can respond. Spoken responses are placed on a FIFO audio queue, allowing the agent loop to continue while ensuring that utterances play completely and in the order they were generated.

\item \textbf{Responsive Local Speech Interaction:} \ludi{} uses a locally deployed speech pipeline optimized for fast human--robot interaction. In our benchmark, once a recorded 10-second English utterance had been captured, the selected four-thread sherpa-onnx configuration \citep{sherpaonnx} transcribed it in 94.7 ms on average, with 99.2 ms P95 processing latency and no transcription errors on the test clips. These measurements reflect STT processing time rather than end-to-end human-perceived response latency.
 More important for the overall system is how speech is integrated into the agent loop: transcripts automatically start new agent turns, speech output runs asynchronously without blocking reasoning or action, queued playback prevents utterances from interrupting one another, and new user input can redirect ongoing navigation or manipulation.

\end{itemize}

\section{Humanoid Navigation: Keypoints and Continuous Planning}

\ludi{} uses a navigation stack that combines conventional localization and keypoint-based navigation with a continuous planner for motions requiring finer humanoid-specific control. Localization combines KISS-ICP LiDAR odometry \citep{vizzo2023kissicp} with an extended Kalman filter, with hyperparameters tuned automatically using Optuna \citep{akiba2019optuna}. For most household tasks, the VLM does not reason directly over metric coordinates. Instead, it navigates through a static map of named keypoints associated with semantically meaningful locations such as rooms, desks, tables, and work areas. The agent can query the available keypoints and then issue a navigation command using the location name. This provides a simple interface between high-level language reasoning and the lower-level navigation stack while keeping metric planning outside the VLM. Before invoking a manipulation skill, the agent uses the \tooltag{reachability} tool to determine whether the target object is within arm's reach. This creates a natural handoff between navigation and manipulation: navigation brings the robot to a semantically meaningful location, while the reachability check determines whether the object can be acted on from the current pose.

When finer local motion is needed, the system invokes an orientation-aware continuous planner tailored to the G1 humanoid. The planner represents position and orientation with quintic spline trajectories, with heading planned independently of direction of travel. It uses an orientation-aware multi-circle footprint to account for the robot’s body geometry during turning, strafing, backward motion, and passage through narrow spaces. The same trajectory representation supports planning, control, replanning, and safety checking, allowing the robot to replan smoothly without repeatedly stopping and restarting.

\section{Specialized VLA Manipulation Policies}

\begin{figure}[h]
    \centering
    \includegraphics[width=0.62\linewidth]{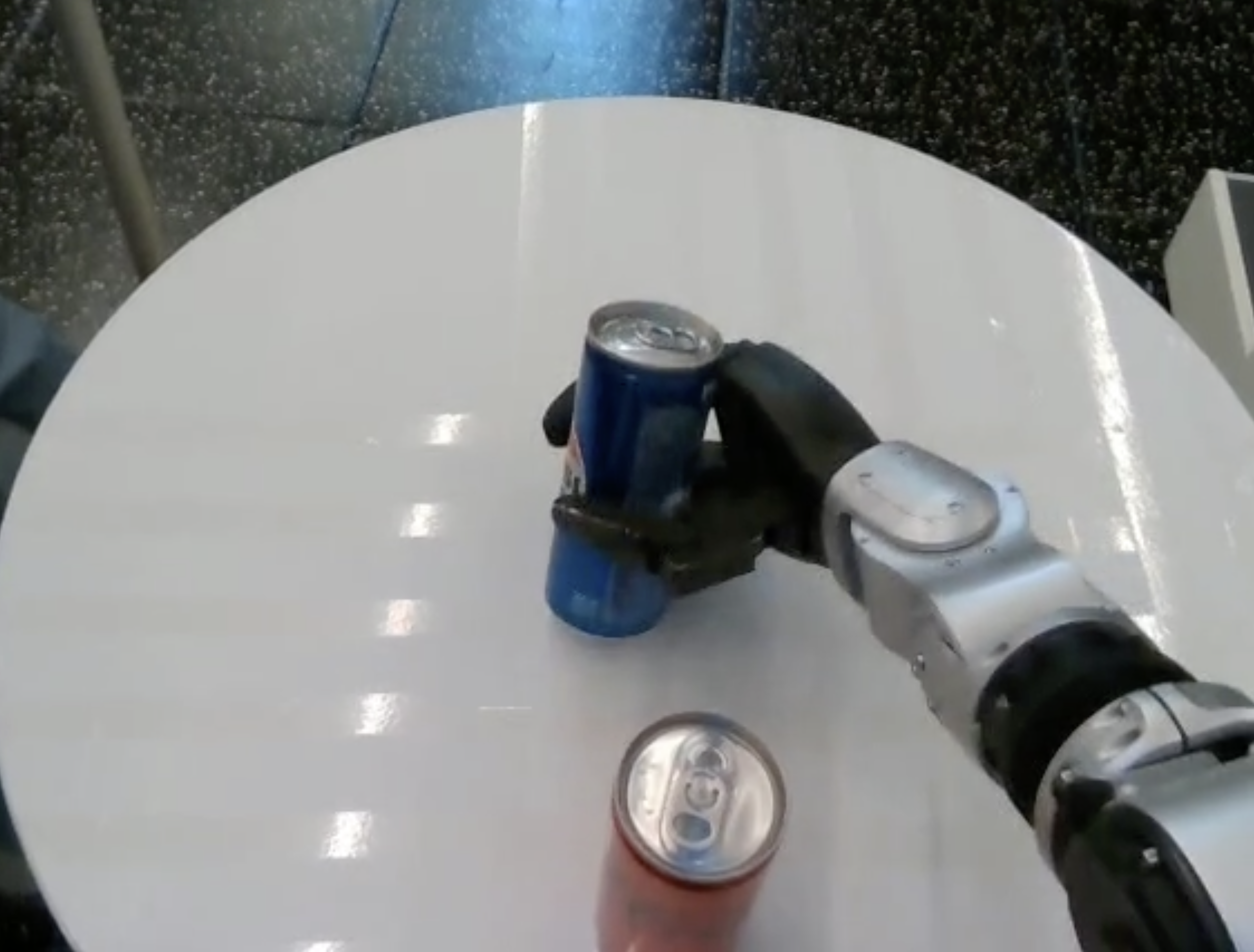}
    \caption{Example of a real-robot pick-and-place task using the learned VLA policy. The Unitree G1, equipped with a Dex-3 hand, grasps the target can from the tabletop using egocentric visual observations.}
    \label{fig:vla_pick}
\end{figure}

Physical manipulation in \ludi{} is carried out by specialized VLA policies invoked as tools by the agent. For the \ludi{} demonstration, we fine-tuned GR00T N1.7 \citep{bjorck2025groot} on the manipulation skills needed in the target home environment, as illustrated in Figure~\ref{fig:vla_pick}.
The VLA setup has four key features:

\begin{itemize}

\item \textbf{Pretrained G1 embodiment:}
We initialized from GR00T N1.7's pretrained Unitree G1 embodiment components (\verb|UNITREE_G1|), including the embodiment-specific state and action representations, and fine-tuned the model directly on our task-specific demonstrations.

\item \textbf{Arm-and-gripper control:}
The VLA controls the robot’s manipulation degrees of freedom only, predicting arm and gripper actions rather than whole-body motion. Navigation and other body motion are handled separately by the robot’s locomotion stack.

\item \textbf{Chunked action prediction:}
The VLA predicts and executes an action horizon of 30 steps synchronously at each inference call. With the robot operating at 30\,Hz, each chunk corresponds to approximately one second of motor commands before the next VLA inference.

\item \textbf{Real-robot fine-tuning:} Fine-tuning data were collected through teleoperation on the physical G1 in household pick-and-place settings similar to those used in the demonstration. This grounds the pretrained model in the robot's actual visual observations, kinematics, and manipulation behavior.

\end{itemize}
These VLA policies serve as learned manipulation skills rather than as the system's primary reasoning component. The VLM interprets the user's request and the broader interaction, decides when manipulation is appropriate, and invokes a VLA with a task instruction. The VLA executes the physical behavior and returns its outcome to the agent when the episode completes or is interrupted.

\paragraph{Calibrated simulation environment.}
Although the final VLA policies were fine-tuned only on real-robot demonstrations, we developed a high-fidelity Isaac Sim model of the G1 \citep{isaacsim} to accelerate policy development and evaluation before physical-robot rollouts. Simulation provided a repeatable testbed for comparing model architectures, training procedures, execution horizons, action representations, and data mixtures, while closed-loop rollouts exposed policies that had low training loss or strong open-loop metrics but failed to complete the task.

To make these comparisons meaningful, we calibrated system timing, camera pose, scene geometry, and motor dynamics separately against recorded behavior of the physical robot. Camera calibration achieved a mask intersection-over-union (IoU) of approximately 0.92, while Dex-3 hand tracking error decreased from 0.132 to 0.043 rad ($\sim$7.6$^\circ$ to 2.5$^\circ$) with fingertip error reduced to at most 13 mm. The fitted dynamics also generalized to 17 held-out episodes from other tasks. We therefore used simulation as a screening tool rather than a replacement for real-robot evaluation: because simulated performance did not always preserve policy rankings on the physical robot, promising policies were ultimately validated through real-world rollouts.

\section{Related Work}

A growing line of work uses large language or vision--language models as high-level decision makers over grounded robot skills. SayCan combines a language model's estimate of an appropriate next action with learned affordance values that reflect what the robot can actually execute \citep{ahn2022saycan}. Inner Monologue closes this loop by returning scene descriptions, success signals, and human feedback to the model so that it can revise its plan \citep{huang2022innermonologue}. Code as Policies instead uses a code-generating language model to compose perception and control APIs directly. Together, these systems illustrate an early agentic pattern in robotics: a general model selects and composes actions, while specialized components execute them and report the results \citep{liang2022codeaspolicies}. More recent systems make this decomposition more explicit. Agentic Robot \citep{yang2025agenticrobot} coordinates a reasoning model, VLA executor, and temporal verifier in a closed planning--execution--verification loop, while VLAs-as-Tools \citep{lei2026vlastools} uses a high-level VLM agent for scene analysis, planning, and recovery, with specialized VLA policies carrying out individual physical subtasks.

A closely related agentic paradigm has been explored with virtual embodiment in PUBG Ally, where an autonomous agent reasons over player input and game state, selects among tools for speech, perception, memory, and action, and iteratively decides what to do next \citep{krafton2026pubgally}. PUBG Ally provides a close precedent for the architecture considered here, but with embodiment in a virtual rather than physical environment.

Other work has focused more directly on the interactive challenges that arise when robots work with people. KnowNo calibrates the uncertainty of a language-model planner so that the robot can ask for help when a request is ambiguous \citep{ren2023knowno}. HELPER-X uses a memory-augmented language model across several interactive embodied domains, including dialogue-based task execution, instruction following, and active question asking \citep{sarch2024helperx}. Related work has also explored language as a mechanism for intervening in ongoing robot behavior: RT-H \citep{belkhale2024rth} supports language-based corrections during policy execution, OLAF \citep{liu2023olaf} uses verbal corrections to refine robot behavior after deployment, and Hi Robot \citep{shi2025hirobot} uses a hierarchical VLM/VLA architecture to incorporate complex instructions and situated user feedback as a task unfolds. These efforts are closely related to \ludi{}'s emphasis on clarification, evolving user intent, and memory across turns. More broadly, recent HRI work shows that LLM-powered robots create interaction requirements distinct from text- and voice-only agents, including stronger expectations for nonverbal behavior \citep{kim2024llmhri}.

In parallel, VLA foundation models such as RT-2 and GR00T learn robot policies conditioned on visual observations and language instructions \citep{zitkovich2023rt2,bjorck2025groot}. 
A particularly close contemporary parallel is Gemini Robotics ER 2, an embodied-reasoning model that can communicate with users, plan and monitor multi-step tasks, and invoke lower-level VLA or navigation capabilities as tools \citep{deepmind2026geminiroboticser2}.

\ludi{} lies at the intersection of these directions but emphasizes training for sustained human--robot interaction. Where several early agentic systems relied primarily on prompting general models, we fine-tuned the VLM on multi-turn traces containing ambiguity, clarification, social dialogue, interruptions, corrections, and tool outcomes. The resulting VLM reasons over transcribed speech, multi-view visual input, interaction history, and prior action outcomes, while a purpose-built harness manages low-latency execution across speech, navigation, and GR00T-based manipulation policies. Rather than treating agentic systems and robot foundation models as competing alternatives, we are advancing a practical agentic system today while using its interaction traces to train a more deeply integrated foundation model for robots and people.

\section{The Limits (and Promise) of Agentic Robotics}

\ludi{} is modular at the execution layer but integrated at the interaction layer. A central VLM reasons over the shared interaction context and decides whether the robot should speak, clarify, wait, navigate, or manipulate, while the harness manages the interaction loop and specialized components execute the selected actions. This architecture enables real-time clarification, handling of interruptions, memory-informed behavior, and coordination between conversation and physical action.

But this integration is operational rather than fully learned. Speech, memory, navigation, and manipulation remain distributed across separate components, and coordinating them through model calls and tools can introduce context loss, latency, brittle handoffs, and fragmented behavior. Although the agent reasons over a shared interaction history, the system does not learn a unified representation of the user, the environment, and the robot's evolving physical state.

We are therefore pursuing two complementary paths. First, we will continue advancing the agentic architecture, improving the VLM, harness, memory, tools, and interaction loop to make Ludi more capable, responsive, and reliable today. Second, we will use the multimodal interaction traces produced by \ludi{}, connecting perception, speech, human feedback, decisions, and physical action, to develop a more deeply integrated foundation model for robots and people. 

\section{Toward \ludione: A Foundation Model for Robots and People}

The next step in this effort is \ludione{}, a foundation model for robots and people that aims to more deeply integrate perception, dialogue, memory, reasoning, and control. Rather than treating conversation as an interface layered on top of robot behavior, \ludione{} aims to ground language and physical action in a shared representation, allowing each to continually inform the other.

Such a model must jointly represent the evolving physical state of the robot and environment, the history and intent of the user, and the relationship between language and action. It should understand how these quantities evolve together over time: what has already happened, what the robot is currently trying to accomplish, what the user expects, and how new observations or utterances change the appropriate course of action. This representation should support not only choosing what to do next, but recognizing when an action is no longer appropriate, when additional information is needed, and when to explain, clarify, revise, pause, or stop as the interaction unfolds.

\ludi{} is therefore both a working embodied agent and a testbed for developing these capabilities. By exposing the limits of current architectures and collecting traces of human--robot interaction, it helps reveal what a future foundation model must learn to represent. Ultimately, the goal is a robot whose physical and social intelligence are deeply integrated: one that understands people, communicates naturally, and acts as a collaborative partner in one coherent, ongoing interaction.

\section{Authorship}

\ludi{} was developed by the following contributors at Ludo Robotics:
\smallskip

\noindent Wooseong Chung,
William Cong$^{*,2}$,
Jakub Dworakowski,
Ethan Ewer$^{*, 4}$,
Tri Wahyu Guntara$^{4}$,
Yeonwoo Jeong,
Tianchong Jiang$^{*,3}$,
Chaewon Kim$^{4}$,
Hyunseo Kim$^{*}$,
Jinwoo Kim$^{4}$,
Jinyeon Kim$^{4}$,
Yea-Seul Kim,
Jack Kunde$^{*}$,
Kangwook Lee$^{4}$,
Sangheon Lee$^{*}$,
Robert Nowak$^{2}$,
and Junha Roh.

\smallskip
{\footnotesize
\noindent
$^{*}$Intern.
$^{2}$Also University of Wisconsin--Madison.
$^{3}$Also Toyota Technological Institute at Chicago.
$^{4}$Also KRAFTON.\par
}

\bibliography{refs}

@article{ahn2022saycan,
  title   = {Do As {I} Can, Not As {I} Say: Grounding Language in Robotic Affordances},
  author  = {Ahn, Michael and Brohan, Anthony and Brown, Noah and others},
  journal = {arXiv preprint arXiv:2204.01691},
  year    = {2022}
}

@article{huang2022innermonologue,
  title   = {Inner Monologue: Embodied Reasoning through Planning with Language Models},
  author  = {Huang, Wenlong and Xia, Fei and Xiao, Ted and others},
  journal = {arXiv preprint arXiv:2207.05608},
  year    = {2022}
}

@misc{krafton2026pubgally,
  author       = {{KRAFTON AI}},
  title        = {From Workflow-Based SLM to Autonomous Agent: Evolving PUBG Ally's Architecture},
  year         = {2026},
  month        = apr,
  howpublished = {KRAFTON AI Blog},
  url          = {https://www.krafton.ai/blog/posts/2026-04-15-pubg_ally_nemotron/pubg-ally-nemotron-en.html}
}

@misc{qwen35,
  title  = {{Qwen3.5}: Towards Native Multimodal Agents},
  author = {{Qwen Team}},
  month  = {February},
  year   = {2026},
  url    = {https://qwen.ai/blog?id=qwen3.5}
}

@inproceedings{goyal2017vqa,
  author    = {Goyal, Yash and Khot, Tejas and Summers-Stay, Douglas and
               Batra, Dhruv and Parikh, Devi},
  title     = {Making the {V} in {VQA} Matter: Elevating the Role of Image
               Understanding in Visual Question Answering},
  booktitle = {Proceedings of the IEEE Conference on Computer Vision and
               Pattern Recognition},
  year      = {2017},
  pages     = {6325--6334}
}

@inproceedings{hu2022lora,
title={Lo{RA}: Low-Rank Adaptation of Large Language Models},
author={Edward J Hu and yelong shen and Phillip Wallis and Zeyuan Allen-Zhu and Yuanzhi Li and Shean Wang and Lu Wang and Weizhu Chen},
booktitle={International Conference on Learning Representations},
year={2022},
url={https://openreview.net/forum?id=nZeVKeeFYf9}
}

@article{deitke2022,
  title={{P}roc{THOR}: {L}arge-Scale Embodied {AI} Using Procedural Generation},
  author={Deitke, Matt and VanderBilt, Eli and Herrasti, Alvaro and Weihs, Luca and Ehsani, Kiana and Salvador, Jordi and Han, Winson and Kolve, Eric and Kembhavi, Aniruddha and Mottaghi, Roozbeh},
  journal={Advances in neural information processing systems},
  volume={35},
  pages={5982--5994},
  year={2022}
}

@InProceedings{hudson2019gqa,
  author    = {Hudson, Drew A. and Manning, Christopher D.},
  title     = {{GQA: A} New Dataset for Real-World Visual Reasoning and Compositional Question Answering},
  booktitle = {Proceedings of the IEEE/CVF Conference on Computer Vision and Pattern Recognition (CVPR)},
  month     = {June},
  year      = {2019},
  pages     = {6700--6709}
}

@inproceedings{yue2024mmmu,
  author    = {Yue, Xiang and Ni, Yuansheng and Zhang, Kai and Zheng, Tianyu
               and Liu, Ruoqi and Ge, Zhang and Stevens, Samuel and Jiang,
               Dongfu and Ren, Weiming and Sun, Yuxuan and others},
  title     = {{MMMU}: A Massive Multi-discipline Multimodal Understanding
               and Reasoning Benchmark for Expert {AGI}},
  booktitle = {Proceedings of the IEEE/CVF Conference on Computer Vision
               and Pattern Recognition},
  year      = {2024},
  pages     = {9556--9567}
}

@misc{sherpaonnx,
  author       = {{The k2-fsa Team}},
  title        = {Sherpa-onnx: Speech-to-text, text-to-speech, and speaker verification using Next-gen Kaldi and ONNX Runtime},
  year         = {2023},
  publisher    = {GitHub},
  journal      = {GitHub Repository},
  howpublished = {\url{https://github.com/k2-fsa/sherpa-onnx}}
}

@article{vizzo2023kissicp,
  author  = {Vizzo, Ignacio and Guadagnino, Tiziano and Mersch, Benedikt and
             Wiesmann, Louis and Behley, Jens and Stachniss, Cyrill},
  title   = {{KISS-ICP}: In Defense of Point-to-Point {ICP} -- Simple, Accurate,
             and Robust Registration If Done the Right Way},
  journal = {IEEE Robotics and Automation Letters},
  volume  = {8},
  number  = {2},
  pages   = {1029--1036},
  year    = {2023},
  doi     = {10.1109/LRA.2023.3236571}
}

@inproceedings{shi2025hirobot,
  author    = {Shi, Lucy Xiaoyang and Ichter, Brian and Equi, Michael Robert
               and Ke, Liyiming and Pertsch, Karl and Vuong, Quan and
               Tanner, James and Walling, Anna and Wang, Haohuan and
               Fusai, Niccolo and Li-Bell, Adrian and Driess, Danny and
               Groom, Lachy and Levine, Sergey and Finn, Chelsea},
  title     = {Hi Robot: Open-Ended Instruction Following with
               Hierarchical Vision-Language-Action Models},
  booktitle = {Proceedings of the 42nd International Conference on Machine Learning},
  series    = {Proceedings of Machine Learning Research},
  volume    = {267},
  pages     = {54919--54933},
  year      = {2025}
}

@article{yang2025agenticrobot,
  title={Agentic {R}obot: {A} brain-inspired framework for vision-language-action models in embodied agents},
  author={Yang, Zhejian and Chen, Yongchao and Zhou, Xueyang and Yan, Jiangyue and Song, Dingjie and Liu, Yinuo and Li, Yuting and Zhang, Yu and Zhou, Pan and Chen, Hechang and others},
  journal={arXiv preprint arXiv:2505.23450},
  year={2025}
}

@article{lei2026vlastools,
  title={Towards Long-horizon Embodied Agents with Tool-Aligned Vision-Language-Action Models},
  author={Lei, Zixing and Liu, Changxing and Xiong, Yichen and Xiong, Minhao and Ding, Yuanzhuo and Zhang, Zhipeng and Li, Weixin and Chen, Siheng},
  journal={arXiv preprint arXiv:2605.13119},
  year={2026}
}

@article{driess2023palme,
  title   = {{PaLM-E}: An Embodied Multimodal Language Model},
  author  = {Driess, Danny and Xia, Fei and Sajjadi, Mehdi S. M. and Lynch, Corey
             and Chowdhery, Aakanksha and Ichter, Brian and Wahid, Ayzaan
             and Tompson, Jonathan and Vuong, Quan and Yu, Tianhe and Huang, Wenlong
             and Chebotar, Yevgen and Sermanet, Pierre and Duckworth, Daniel
             and Levine, Sergey and Vanhoucke, Vincent and Hausman, Karol
             and Toussaint, Marc and Greff, Klaus and Zeng, Andy and Mordatch, Igor
             and Florence, Pete},
  journal = {arXiv preprint arXiv:2303.03378},
  year    = {2023}
}

@article{cheng2024navila,
  title   = {{NaVILA}: Legged Robot Vision-Language-Action Model for Navigation},
  author  = {Cheng, An-Chieh and Ji, Yandong and Yang, Zhaojing and Gongye, Zaitian
             and Zou, Xueyan and Kautz, Jan and B{\i}y{\i}k, Erdem
             and Yin, Hongxu and Liu, Sifei and Wang, Xiaolong},
  journal = {arXiv preprint arXiv:2412.04453},
  year    = {2024}
}

@inproceedings{zitkovich2023rt2,
  author    = {Zitkovich, Brianna and Yu, Tianhe and Xu, Sichun and others},
  title     = {{RT-2}: Vision-Language-Action Models Transfer Web Knowledge to Robotic Control},
  booktitle = {Proceedings of the 7th Conference on Robot Learning},
  series    = {Proceedings of Machine Learning Research},
  volume    = {229},
  pages     = {2165--2183},
  year      = {2023},
  publisher = {PMLR}
}

@inproceedings{kim2025openvla,
  author    = {Kim, Moo Jin and Pertsch, Karl and Karamcheti, Siddharth and
               Xiao, Ted and Balakrishna, Ashwin and Nair, Suraj and Rafailov,
               Rafael and Foster, Ethan P. and Sanketi, Pannag R. and Vuong,
               Quan and Kollar, Thomas and Burchfiel, Benjamin and Tedrake,
               Russ and Sadigh, Dorsa and Levine, Sergey and Liang, Percy and
               Finn, Chelsea},
  title     = {OpenVLA: An Open-Source Vision-Language-Action Model},
  booktitle = {Proceedings of the 8th Conference on Robot Learning},
  series    = {Proceedings of Machine Learning Research},
  volume    = {270},
  pages     = {2679--2713},
  year      = {2024},
  publisher = {PMLR}
}

@article{zhang2026dreamvla,
  title={{DreamVLA: A vision-language-action model dreamed with comprehensive world knowledge}},
  author={Zhang, Wenyao and Liu, Hongsi and Qi, Zekun and Wang, Yunnan and Yu, Xinqiang and Zhang, Jiazhao and Dong, Runpei and He, Jiawei and Wang, He and Zhang, Zhizheng and others},
  journal={Advances in Neural Information Processing Systems},
  volume={38},
  pages={24195--24228},
  year={2025}
}

@article{dreamzero2026,
  title   = {World Action Models are Zero-shot Policies},
  author  = {Ye, Seonghyeon and Ge, Yunhao and Zheng, Kaiyuan and others},
  journal = {arXiv preprint arXiv:2602.15922},
  year    = {2026}
}

@article{zhen2024threedvla,
  title   = {{3D-VLA}: A 3D Vision-Language-Action Generative World Model},
  author  = {Zhen, Haoyu and Qiu, Xiaowen and Chen, Peihao and Yang, Jincheng
             and Yan, Xin and Du, Yilun and Hong, Yining and Gan, Chuang},
  journal = {arXiv preprint arXiv:2403.09631},
  year    = {2024}
}

@article{black2024pi0,
  title   = {$\pi_0$: A Vision-Language-Action Flow Model for General Robot Control},
  author  = {Black, Kevin and Brown, Noah and Driess, Danny and Esmail, Adnan
             and Equi, Michael and Finn, Chelsea and Fusai, Niccolo and Groom, Lachy
             and Hausman, Karol and Ichter, Brian and Jakubczak, Szymon
             and Jones, Tim and Ke, Liyiming and Levine, Sergey and Li-Bell, Adrian
             and Mothukuri, Mohith and Nair, Suraj and Pertsch, Karl
             and Shi, Lucy Xiaoyang and Tanner, James and Vuong, Quan
             and Walling, Anna and Wang, Haohuan and Zhilinsky, Ury},
  journal = {arXiv preprint arXiv:2410.24164},
  year    = {2024}
}

@misc{isaacsim,
  author       = {{NVIDIA Isaac Sim Project Developers}},
  title        = {{NVIDIA Isaac Sim: An Open-Source Reference Framework for Robotics Simulation, Testing, and Synthetic Data Generation}},
  url          = {https://github.com/isaac-sim/IsaacSim},
  year         = {2026},
  note         = {Built on NVIDIA Omniverse}
}

@inproceedings{akiba2019optuna,
  author    = {Akiba, Takuya and Sano, Shotaro and Yanase, Toshihiko and
               Ohta, Takeru and Koyama, Masanori},
  title     = {Optuna: A Next-generation Hyperparameter Optimization Framework},
  booktitle = {Proceedings of the 25th ACM SIGKDD International Conference
               on Knowledge Discovery \& Data Mining},
  pages     = {2623--2631},
  year      = {2019},
  doi       = {10.1145/3292500.3330701}
}

@inproceedings{belkhale2024rth,
  author    = {Belkhale, Suneel and Ding, Tianli and Xiao, Ted and
               Sermanet, Pierre and Vuong, Quan and Tompson, Jonathan and
               Chebotar, Yevgen and Dwibedi, Debidatta and Sadigh, Dorsa},
  title     = {{RT-H}: Action Hierarchies using Language},
  booktitle = {Proceedings of Robotics: Science and Systems},
  year      = {2024},
  address   = {Delft, Netherlands},
  month     = jul,
  doi       = {10.15607/RSS.2024.XX.049}
}

@article{liu2023olaf,
  title={Interactive robot learning from verbal correction},
  author={Liu, Huihan and Chen, Alice and Zhu, Yuke and Swaminathan, Adith and Kolobov, Andrey and Cheng, Ching-An},
  journal={arXiv preprint arXiv:2310.17555},
  year={2023}
}

@inproceedings{kim2024llmhri,
  author    = {Kim, Callie Y. and Lee, Christine P. and Mutlu, Bilge},
  title     = {Understanding Large-Language Model (LLM)-powered Human-Robot Interaction},
  booktitle = {Proceedings of the 2024 ACM/IEEE International Conference on Human-Robot Interaction},
  pages     = {371--380},
  year      = {2024},
  doi       = {10.1145/3610977.3634966}
}

@misc{deepmind2026geminiroboticser2,
  author = {Hansen, Steven and Xu, Peng},
  title  = {Introducing {G}emini {R}obotics {ER} 2},
  year   = {2026},
  month  = jul,
  note   = {Google DeepMind}
}

@article{liang2022codeaspolicies,
  title   = {Code as Policies: Language Model Programs for Embodied Control},
  author  = {Liang, Jacky and Huang, Wenlong and Xia, Fei and Xu, Peng and
             Hausman, Karol and Ichter, Brian and Florence, Pete and Zeng, Andy},
  journal = {arXiv preprint arXiv:2209.07753},
  year    = {2022}
}

@article{ren2023knowno,
  title   = {Robots That Ask For Help: Uncertainty Alignment for Large Language Model Planners},
  author  = {Ren, Allen Z. and Dixit, Anushri and Bodrova, Alexandra and others},
  journal = {arXiv preprint arXiv:2307.01928},
  year    = {2023}
}

@article{sarch2024helperx,
  title   = {{HELPER-X}: A Unified Instructable Embodied Agent to Tackle Four
             Interactive Vision-Language Domains with Memory-Augmented Language Models},
  author  = {Sarch, Gabriel and Somani, Sahil and Kapoor, Raghav and
             Tarr, Michael J. and Fragkiadaki, Katerina},
  journal = {arXiv preprint arXiv:2404.19065},
  year    = {2024}
}

@inproceedings{radford2021clip,
  title={Learning transferable visual models from natural language supervision},
  author={Radford, Alec and Kim, Jong Wook and Hallacy, Chris and Ramesh, Aditya and Goh, Gabriel and Agarwal, Sandhini and Sastry, Girish and Askell, Amanda and Mishkin, Pamela and Clark, Jack and others},
  booktitle={International {C}onference on {M}achine {L}earning},
  pages={8748--8763},
  year={2021},
  organization={PMLR}
}

@article{alayrac2022flamingo,
  title={Flamingo: a Visual Language Model for Few-Shot Learning},
  author={Alayrac, Jean-Baptiste and Donahue, Jeff and Luc, Pauline and Miech, Antoine and Barr, Iain and Hasson, Yana and Lenc, Karel and Mensch, Arthur and Millican, Katie and Reynolds, Malcolm and others},
  journal={Advances in Neural Information Processing Systems (NeurIPS)},
  volume={35},
  pages={23716--23736},
  year={2022}
}

@article{liu2023llava,
  title={Visual instruction tuning},
  author={Liu, Haotian and Li, Chunyuan and Wu, Qingyang and Lee, Yong Jae},
  journal={Advances in Neural Information Processing Systems},
  volume={36},
  pages={34892--34916},
  year={2023}
}

@article{bjorck2025groot,
  title={GR00T N1: An Open Foundation Model for Generalist Humanoid Robots},
  author={Bjorck, Johan and Casta{\~n}eda, Fernando and Cherniadev, Nikita and Da, Xingye and Ding, Runyu and Fan, Linxi and Fang, Yu and Fox, Dieter and Hu, Fengyuan and Huang, Spencer and others},
  journal={arXiv preprint arXiv:2503.14734},
  year={2025}
}

@article{ai2thor,
  title={{AI2-THOR}: An interactive 3d environment for visual {AI}},
  author={Kolve, Eric and Mottaghi, Roozbeh and Han, Winson and VanderBilt, Eli and Weihs, Luca and Herrasti, Alvaro and Deitke, Matt and Ehsani, Kiana and Gordon, Daniel and Zhu, Yuke and others},
  journal={arXiv preprint arXiv:1712.05474},
  year={2017}
}

@inproceedings{agrawal2026gepa,
  title={{GEPA}: {R}eflective prompt evolution can outperform reinforcement learning},
  author={Agrawal, Lakshya A and Tan, Shangyin and Soylu, Dilara and Ziems, Noah and Khare, Rishi and Opsahl-Ong, Krista and Singhvi, Arnav and Shandilya, Herumb and Ryan, Michael J and Jiang, Meng and others},
  booktitle={International Conference on Learning Representations},
  volume={2026},
  pages={8479--8565},
  year={2026}
}

\end{document}